\documentclass[letterpaper, 10 pt, conference]{ieeeconf}  

\IEEEoverridecommandlockouts                              

\usepackage{graphics} 
\usepackage{epsfig} 
\usepackage{mathptmx} 
\usepackage{times} 
\usepackage{amsmath} 
\usepackage{amssymb}  
\usepackage{balance}
\usepackage{multirow}
\usepackage{multicol}
\usepackage{makecell}
\usepackage{hyperref}
\usepackage{array}
\usepackage{cite}
\usepackage{url}
\usepackage{xcolor}
\usepackage{subcaption}
\usepackage{arydshln}
\usepackage{array}
\newcolumntype{C}[1]{{\centering\arraybackslash}p{#1}}

\title{\LARGE \bf
4D mmWave Radar for Sensing Enhancement in Adverse Environments: Advances and Challenges}

\author{Xiangyuan Peng$^{1,2\dagger}$  Miao Tang$^{3\dagger}$ Huawei Sun$^{1,2}$ Kay Bierzynski$^{2}$ Lorenzo Servadei$^{1}$ and Robert Wille$^{1}$
\thanks{$^{1}$Technical University of Munich, Munich, Germany}%
\thanks{$^{2}$Infineon Technologies AG, Neubiberg,
        Germany
   }%
   \thanks{$^{3}$China University of Geosciences, Wuhan, China
   }%
   \thanks{$^{\dagger}$Xiangyuan Peng and Miao Tang contribute equally to this work.}
}

\begin{document}

\maketitle
\thispagestyle{empty}
\pagestyle{empty}






\begin{abstract}
Intelligent transportation systems require accurate and reliable sensing. 
However, adverse environments, such as rain, snow, and fog, can significantly degrade the performance of LiDAR and cameras. In contrast, 4D mmWave radar not only provides 3D point clouds and velocity measurements but also maintains robustness in challenging conditions. Recently, research on 4D mmWave radar under adverse environments has been growing, but a comprehensive review is still lacking.
To bridge this gap, this work reviews the current research on 4D mmWave radar under adverse environments. First, we present an overview of existing 4D mmWave radar datasets encompassing diverse weather and lighting scenarios. Subsequently, we analyze existing learning-based methods leveraging 4D mmWave radar to enhance performance according to different adverse conditions. Finally, the challenges and potential future directions are discussed for advancing 4D mmWave radar applications in harsh environments. To the best of our knowledge, this is the first review specifically concentrating on 4D mmWave radar in adverse environments. 
The related studies are listed in: \href{https://github.com/XiangyPeng/4D-mmWave-Radar-in-Adverse-Environments}{https://github.com/4DR-Adv}.

\end{abstract}


\section{Introduction}
\label{sec:intro}
%

Human error remains the primary cause of traffic accidents in today’s transportation systems \cite{segun2024factors}. To address this challenge, modern Intelligent Transportation Systems (ITS) aim to enhance road safety and efficiency through robust autonomous driving. The safety and reliability of autonomous driving heavily rely on accurate sensing and scene understanding.
However, adverse environments severely affect the robustness of ITS \cite{suk2024addressing}. Sun glare, rain, fog, and snow can degrade the sensing performance by introducing noise and occlusions \cite{qi2024geometric}. Raindrops affect road surface conditions by creating driving hazards. According to statistics from the US Department of Transportation, around 11\% of 6 million traffic accidents occur on snowy or icy roads \cite{oh2025assessing}. Furthermore, phenomena such as fog, sandstorms, and smoke also reduce the visibility \cite{9167446}. 

\begin{figure}[t]
    \centering
    \begin{subfigure}[t]{0.49\linewidth}
        \centering
        \includegraphics[width=\linewidth]{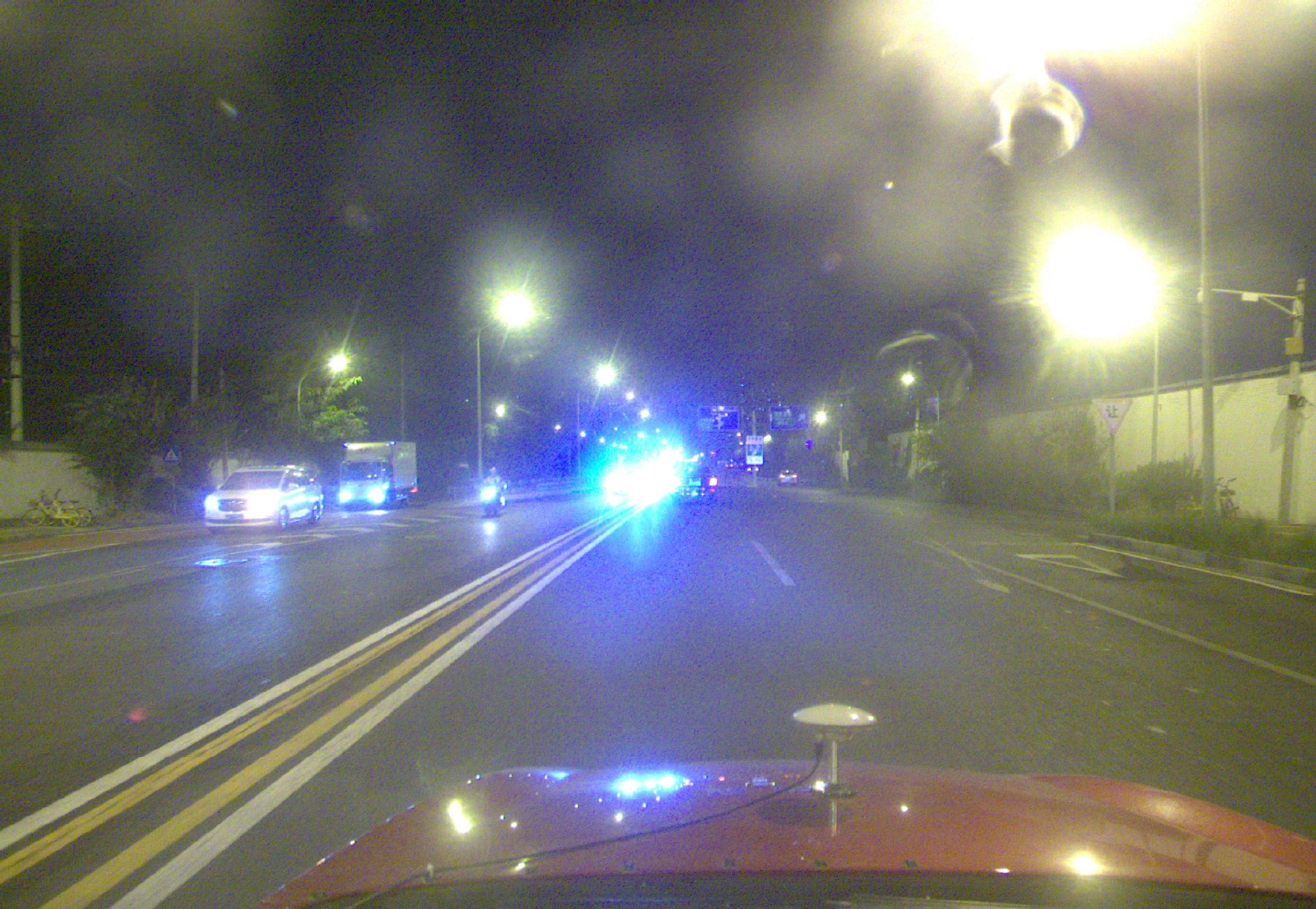}
        \vspace{-5mm}
        \caption{}
        \label{fig:sensors_weather_left}
    \end{subfigure}
    \begin{subfigure}[t]{0.49\linewidth}
        \centering
        \includegraphics[width=\linewidth]{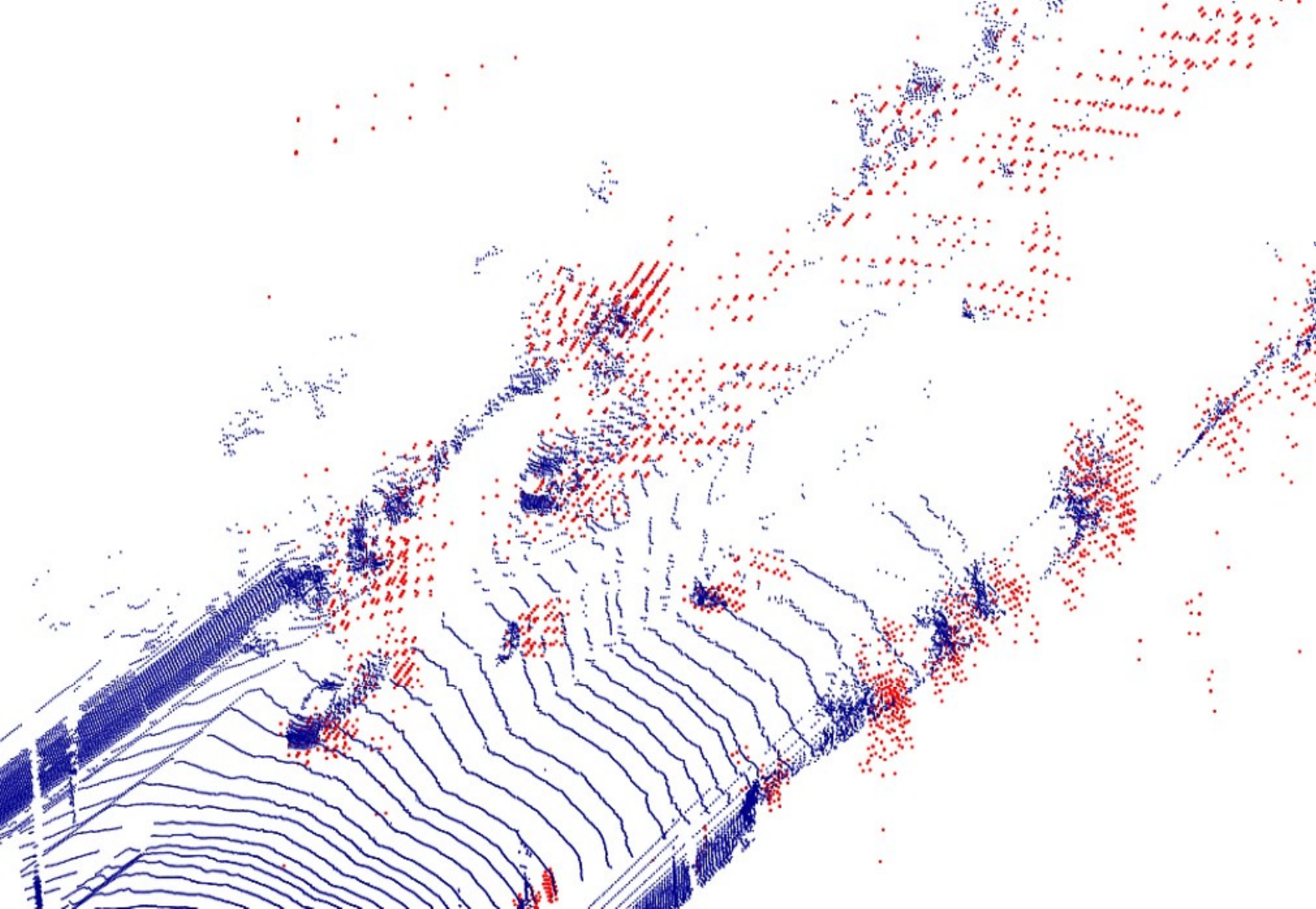}
        \vspace{-5mm}
        \caption{}
        \label{fig:sensors_weather_right}
    \end{subfigure}
    \caption{Visualization of camera, LiDAR and 4D mmWave data in rainy nighttime. The raw data is from the Dual Radar dataset \cite{zhang2023dual}. (a) shows the camera image. (b) illustrates LiDAR point clouds in blue and 4D mmWave radar (Arbe Phoenix) point clouds in red.}
    \label{fig:sensors_weather_combined}
    \vspace{-4mm}
\end{figure}

Modern autonomous driving systems commonly adopt multi-sensor setups. Therefore, an increasing number of public datasets incorporate multiple sensors for perception and SLAM tasks. Fig. \ref{fig:sensors_weather_combined} shows camera, radar, and LiDAR data under rainy nighttime conditions from the Dual Radar dataset. 
Cameras excel in 2D detection by high-resolution RGB images but struggle with occlusion and poor depth perception under challenging lighting and weather \cite{ma2024licrocc}. LiDAR generates 3D point clouds with precise geometric information and is less affected by lighting \cite{jiang2024weather}. However, LiDAR's measurement accuracy and point density deteriorate due to noisy interference of adverse conditions.

\begin{figure}[t] 
    \centering
    \includegraphics[width=0.8\linewidth]{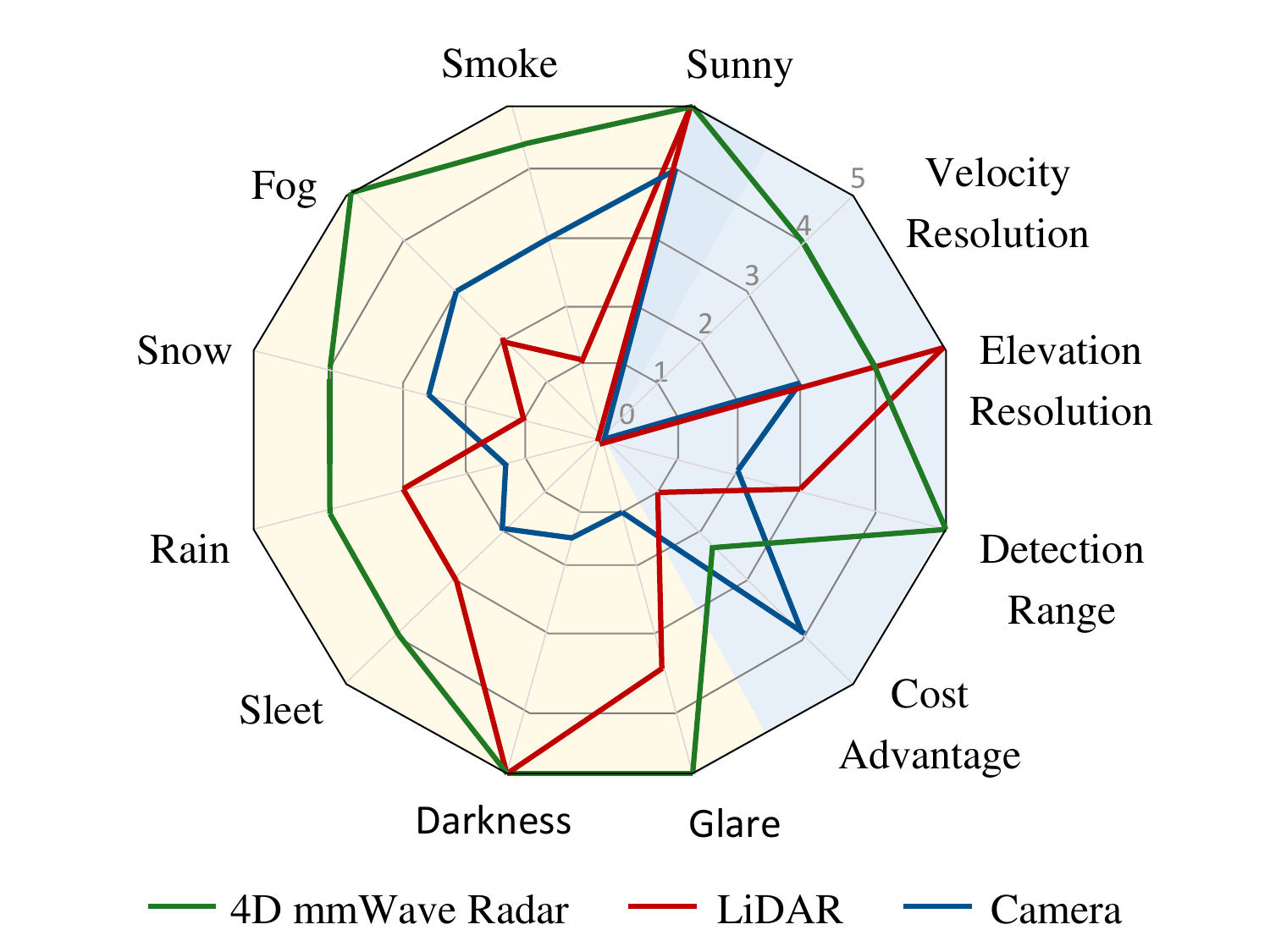}
    \vspace{-1mm}
    \caption{Analysis of 4D mmWave radar, LiDAR, and cameras under adverse environments and their features.} 
    \label{fig: sensors under weather}
    \vspace{-5mm}
\end{figure}

\begin{figure*}[t]
    \centering
    \includegraphics[width=0.95\textwidth]{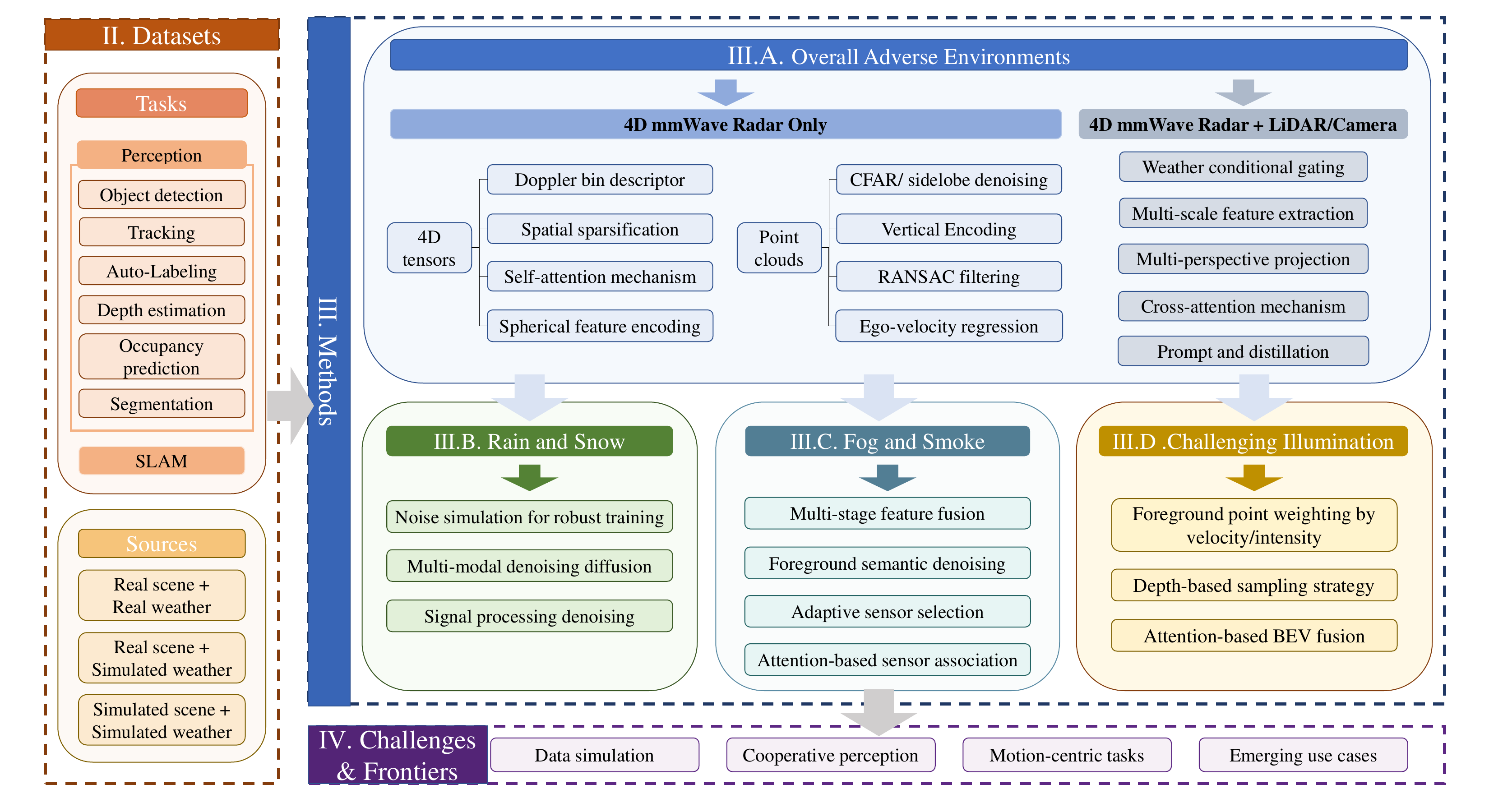}
    \vspace{-3mm}
    \caption{The overall framework of the review.}
    \label{fig: overall}
    \vspace{-6mm}
\end{figure*}
As an alternative, 4D millimeter-wave (mmWave) radar is widely employed due to its compact size, cost-effectiveness, velocity-measuring capability, long detection range, and all-weather adaptability \cite{kong2024survey}. It enhances traditional 3D mmWave radar by adding height measurement, enabling 3D spatial information. 
Compared to other sensors, 4D mmWave radar can penetrate small airborne particles, achieving consistent and reliable operation in challenging environments \cite{fan20244d}.
A detailed comparison of cameras, LiDAR, and 4D mmWave radar across different environments and their features is presented in Fig. \ref{fig: sensors under weather}. The comparison indicates that 4D mmWave radar outperforms LiDAR and cameras in rain, snow, sleet, fog, and smoke conditions, with superior detection range and velocity measurement.
Therefore, 4D mmWave radar-only methods offer a balance of robustness and cost-efficiency. When fusing with other sensors, 4D mmWave radar effectively complements the limitations of cameras and LiDAR, overcoming issues such as short detection range and reduced performance in adverse environments.

Advancements have introduced weather-included 4D mmWave radar datasets and methods \cite{paek2022k, chae2024towards,matuszka2022aimotive}. These studies target different scenes, such as land, waterways, and mines, as well as different applications, like single-agent and cooperative perception. Several reviews have been conducted on 4D mmWave radar technologies. 
The first review of deep learning (DL) -based 4D mmWave radar datasets and their applications was presented in \cite{han20234d}. Subsequent reviews have explored its advantages over other sensors in specific tasks, such as detection and tracking \cite{liu2024framework,fan20244d}.
However, there remains a lack of a dedicated review specifically centering on 4D mmWave radar enhancing the performance under challenging conditions. 
To bridge this gap, this paper comprehensively summarizes 4D mmWave radar in adverse environments, providing valuable insights and a foundation for future research.
The main contributions are as follows:
\begin{itemize}
    \item To the best of our knowledge, this is the first survey comprehensively summarizing the current DL-based 4D mmWave radar datasets, methods, and applications designed for adverse environments.
    \item A detailed overview of 4D mmWave radar datasets under adverse conditions for perception and SLAM is provided, covering diverse weather and illumination.
    \item Existing DL-based models using 4D mmWave radar to enhance performance in adverse environments are analyzed, including general strategies for multiple conditions and specialized techniques for specific weather. 
    \item Challenges and frontiers are discussed for 4D mmWave radar in adverse environments, focusing on simulation realism, cooperative perception, motion-centric tasks, and emerging application cases.
\end{itemize}

The overall framework of this review is shown in Fig. \ref{fig: overall}.
First, public datasets covering various weather and lighting conditions are introduced in section \ref{sec: datasets}. Section \ref{sec: methods} provides a detailed analysis of DL-based research with 4D mmWave radar under different adverse conditions. And section \ref{sec: discussion} discusses the challenges and frontiers for 4D mmWave radar in diverse environments. 

\section{Dataset}
\label{sec: datasets}

Numerous 4D mmWave radar datasets have been developed. Datasets such as Astyx \cite{8904734}, VoD \cite{apalffy2022}, RADIal \cite{Rebut_2022_CVPR}, and ZJOUDset \cite{10401068} focus on 3D object detection based on 4D mmWave radar with synchronized LiDAR, and camera data. 
RaDelft \cite{10731871}, SJTU4D \cite{10173499}, Colorado \cite{kramer2022coloradar}, and DRIO\cite{10207713} provide radar point clouds and multisensor data with ground truth poses for Simultaneous Localization and Mapping (SLAM) tasks. 
Although these datasets have facilitated significant advancements in research under standard environments, they are limited in evaluating performance in more complex and adverse environments.
Therefore, the performance of 4D mmWave radar under adverse conditions has gained increasing attention in recent years \cite{paek2022k}. The current public 4D mmWave radar datasets with adverse environments are summarized in Table \ref{tab: main datasets}.

\setlength\tabcolsep{4.5pt}
\begin{table*}[ht]
\centering
\caption{Publicly available 4D mmWave radar datasets with adverse weather and light conditions.}
\label{tab: main datasets}
\vspace{-2mm}
\begin{tabular}{c|c|c|c|c|c|cccccc|ccc}
\hline
\multirow{2}{*}{Dataset} &\multirow{2}{*}{Year} &\multirow{2}{*}{Modality} &\multirow{2}{*}{Source}&\multirow{2}{*}{Task} &\multirow{2}{*}{Size} &\multicolumn{6}{c|}{Adverse Weather} &\multicolumn{3}{c}{Adverse Illumination} \\
\cline{7-15}
 & & & & & &fog &rain &sleet &snow &gloom &smoke &night &glare &low-light \\
\hline
\multicolumn{15}{c}{Datasets for Perception} \\
\hline
TJ4DRadSet\cite{9922539}&2022 &R,L,C &RT&OD,T  &7.8K&&&&& & &$\checkmark$& & \\
\hline
K-Radar\cite{paek2022k} &2022 &R,L,C&RT &OD,T&35K &$\checkmark$&$\checkmark$&$\checkmark$&$\checkmark$&$\checkmark$ & &$\checkmark$& & \\
\hline
aiMotive\cite{matuszka2022aimotive} &2022&R,L,C &RT  &OD,T &26.5K& &$\checkmark$&& &$\checkmark$ & &$\checkmark$&$\checkmark$&\\
\hline
Dual Radar\cite{zhang2023dual} &2023 &R,L,C &RT  &OD,T&50K & &$\checkmark$& & &$\checkmark$ & &$\checkmark$& &$\checkmark$\\
\hline
L-Radset\cite{10591452} &2024 &R,L,C &RT &OD,T &11.2K&$\checkmark$&$\checkmark$& & &$\checkmark$ & &$\checkmark$& &$\checkmark$ \\
\hline
Bosch Street\cite{armanious2024bosch} &2024 &R,L,C  &RT&OD  &1.3M & &$\checkmark$& & &$\checkmark$ & &$\checkmark$& &$\checkmark$\\
\hline
MAN TruckScenes\cite{fent2024man}&2024 &R,L,C,I &RT &OD,T &30K & &$\checkmark$ & &$\checkmark$ &$\checkmark$ & &$\checkmark$&$\checkmark$ &$\checkmark$\\
\hline
CMD\cite{deng2024cmd} &2024  &R,L,C & RT&OD,T& 10K& & & & & & &$\checkmark$& &$\checkmark$\\
\hline
VOD-Fog\cite{huang2024l4dr} &2024 &R,L,C,I &SIM &OD,T &8.7K &$\checkmark$& & & & & & & & \\
\hline
V2X-Radar\cite{yang2024v2x} &2024 &R,L,C,I &RT&OD  &20K&&$\checkmark$ & &$\checkmark$& & &$\checkmark$ & &$\checkmark$ \\
\hline
V2X-R\cite{V2X-R} &2024 &R,L,C,I &SIM&OD  &37.7K&$\checkmark$& & &$\checkmark$& & & & & \\
\hline
ZJU-Multispectrum\cite{10623522} &2024&R,C$_t$ &RT  &DE &21.5K & & & &&&$\checkmark$&$\checkmark$ &&\\
\hline
OmniHD-Scenes\cite{zheng2024omnihd} &2024&R,L,C &RT  &OP,OD,T &450K& &$\checkmark$ & & & $\checkmark$& &$\checkmark$& &$\checkmark$\\
\hline
WaterScenes\cite{yao2024waterscenes} &2024 &R,C,I&RT &OD,S&54K &$\checkmark$&$\checkmark$& &$\checkmark$&$\checkmark$  & &$\checkmark$&$\checkmark$ &$\checkmark$ \\
\hline
\multicolumn{15}{c}{Datasets for SLAM} \\
\hline
MSC RAD4R\cite{10225273} &2023  &R,L,C,I &RT&SLAM &90.8K & & & &$\checkmark$& &$\checkmark$ &$\checkmark$& & \\
\hline
NTU4DRadLM\cite{10422606} &2023 &R,L,C,C$_t$,I &RT&SLAM &61K & &$\checkmark$ & & &$\checkmark$ & &$\checkmark$& &\\
\hline
DIDLM\cite{gong2024didlm} &2024 &R,L,C,C$_i$,I & RT&SLAM &- & &$\checkmark$& &$\checkmark$& & &$\checkmark$& & \\
\hline
MINE4DRAD\cite{10684300} &2024  &R,L,C,I &RT&SLAM&- & & & & &$\checkmark$  &$\checkmark$&$\checkmark$& &\\
\hline
SNAIL Radar\cite{huai2024snail} &2024 &R,L,C,I &RT&SLAM &- & & $\checkmark$ & &  & &&$\checkmark$& &$\checkmark$\\
\hline
HeRCULES\cite{kim2025hercules} &2025 &R,L,C,I &RT&SLAM &- & & $\checkmark$ & & $\checkmark$ & &&$\checkmark$& &$\checkmark$ \\
\hline
NavINST \cite{de2025navinst} &2025  &R,L,C,I &RT&SLAM&- & &  & &  & &&$\checkmark$& & \\
\hline
\end{tabular}
\\[1pt]
\scriptsize{OD, T, OP, S, and DE denote object detection, tracking, occupancy prediction, segmentation, and depth estimation. RT indicates real-world data, while SIM means simulated data. R, L, C, and I represent 4D mmWave radar, LiDAR, camera, and IMU. C$_i$ is the infrared camera, and C$_t$ is the thermal camera. Gloom includes overcast and cloud scenarios. Low-light indicates the dawn and dusk periods.}
\vspace{-6mm}
\end{table*}

\subsection{Datasets for perception}
 
TJ4DRadSet \cite{9922539} comprises 7757 frames of 4D mmWave radar, LiDAR, and camera data for 3D object detection and tracking. It considers diverse driving scenarios such as urban roads, elevated highways, and industrial zones with different lightness, making it a foundation for 4D mmWave radar perception under varying light and driving conditions.

Beyond driving conditions and lighting, to leverage the advantages of the 4D mmWave radar in adverse weather,  K-Radar \cite{paek2022k} is collected as the first large-scale 4D mmWave radar dataset specifically focused on adverse weather. It includes 35K frames of 4D mmWave radar tensors (Doppler, range, azimuth, and elevation) and point clouds. The dataset encompasses a wide range of weather conditions, including normal weather, overcast, fog, rain, sleet, and snow within a middle detection range of 120m.

However, the K-Radar dataset \cite{paek2022k} lacks long-range 4D mmWave radar point clouds. In contrast, the aiMotive dataset \cite{matuszka2022aimotive} provides long-range perception up to 200 meters. Similarly, the Dual Radar dataset \cite{zhang2023dual} incorporates two 4D mmWave radar systems (Arbe Phoenix and ARSS48 RDI), supporting both middle- and long-range detection. The use of different radar systems enables a comparative analysis of various 4D mmWave radar designs under identical scenes. Added to this, the L-RadSet dataset \cite{10591452} extends long-range detection capabilities to 220 meters. While all three datasets cover diverse weather scenarios, including sunny, rainy, cloudy, and nighttime conditions, aiMotive \cite{matuszka2022aimotive} also captures a small number of glare scenes, and L-RadSet additionally features foggy weather data. 

To further enhance the data scale, the Bosch Street dataset \cite{armanious2024bosch} provides 1.3M frames of synchronized data across 13.6K different scenes within a long distance of 200m with a 360° field of view. 
To advance truck-specific autonomous perception, MAN TruckScenes \cite{fent2024man} provides a 4D mmWave radar dataset with a detection range of up to 230 meters. The dataset is collected under diverse weather and lighting conditions, filling a critical gap in tracking for heavy vehicles.

Cross-sensor domain adaptation can mitigate sensor-based disparities. CMD \cite{deng2025cmd} is the first dataset designed for cross-sensor domain adaptation, incorporating data from 4D mmWave radar, cameras, and multiple LiDARs with different beam configurations in diverse lighting conditions. It considers the density, intensity, and geometry across different sensors, supporting research on cross-sensor adaptation.

Collecting real-world data under adverse conditions is challenging. Therefore, simulation has become a practical alternative. Building on the real-world VoD dataset \cite{apalffy2022}, the work of \cite{huang2024l4dr} simulated different levels of fog to LiDAR point clouds while leaving 4D mmWave radar point clouds unchanged. This extension, known as the VoD-Fog dataset, enables research on using 4D mmWave radar-LiDAR fusion for 3D detection and tracking in foggy conditions.

Compared to single-agent perception, cooperative perception is crucial for extending range and overcoming occlusions. However, 4D mmWave radar datasets in the vehicle-to-everything (V2X) domain remain scarce. Therefore, V2X-Radar \cite{yang2024v2x} introduces the first real-world V2X 4D mmWave radar dataset for 3D object detection, covering sunny and rainy scenarios. It includes subsets for cooperative, roadside, and single-vehicle perception. Complementing this, V2X-R \cite{V2X-R} is the first simulated V2X dataset featuring 4D mmWave radar, LiDAR, and cameras via CARLA \cite{dosovitskiy2017carla}. Through simulation, fog and snow weather are incorporated. 
The diffusion model of V2X-R \cite{V2X-R} is utilized for the first time to denoise LiDAR data using 4D mmWave radar features.

Beyond detection and tracking, some 4D mmWave radar datasets tackle other perception tasks. ZJU-Multispectrum \cite{10623522} focuses on depth estimation using 4D mmWave radar and thermal camera. It incorporates artificial smoke scenarios to evaluate interference handling. OmniHD-Scenes \cite{zheng2024omnihd} targets occupancy prediction alongside detection and tracking. The data is collected in sunny, cloudy, and rainy conditions during both day and night.

In addition to the road-based dataset, WaterScenes \cite{yao2024waterscenes} is the first 4D mmWave radar-camera dataset for waterway autonomous navigation. It supports object detection, instance, and semantic segmentation, with 2D boxes and pixel-level labels for images, as well as 3D point labels. It covers diverse weather and lighting conditions in waterway environments.

\subsection{Datasets for SLAM}
Due to the robustness under various conditions, 4D mmWave radar is also increasingly applied to SLAM applications with challenging environments \cite{10225273}.

For large-scale SLAM in urban environments, MSC RAD4R \cite{10225273} features 4D mmWave radar with a long detection range of up to 400m, complemented by odometry sensors. The dataset includes scenarios of normal and snowy weather, along with artificial smoke.

NTU4DRadLM \cite{10422606} is a 4D mmWave radar dataset collected by both robotic and vehicular platforms across various conditions. Derived from it, NTU4DPR \cite{10655664} serves as a benchmark for place recognition with 4D mmWave radar data. DIDLM \cite{gong2024didlm} integrates infrared cameras, depth cameras, LiDAR, and 4D mmWave radar for 3D mapping. It spans indoor and outdoor scenarios, covering adverse weather, including rain and snow.
SNAIL Radar \cite{huai2024snail} is captured by three different platforms, encompassing conditions of rainy days and nights.

HeRCULES \cite{kim2025hercules} is the first dataset to integrate both 4D radar and spinning radar. It encompasses various weather and lighting conditions with a large number of dynamic objects. NavINST \cite{de2025navinst} incorporates four 4D mmWave radars and 1 Doppler radar, generating unified point clouds with 360° coverage under both day and night.

MINE4DRAD \cite{10684300} emerges as the first 4D mmWave radar dataset for open-pit mines, focusing on detecting static obstacles for haulage. The 4D mmWave radar point clouds cover a wide range of dusty and smoky mining scenarios, including loading sites, dumping sites, haulage maintenance areas, and commute roads. It offers valuable data for addressing the specific challenges of open-pit mine environments.

The aforementioned 4D mmWave radar datasets address various tasks for perception and SLAM. By encompassing adverse weather conditions and lighting scenarios through real-world data collection and simulation, these datasets establish a foundation for advancing 4D mmWave radar research in challenging environments.

\section{Method}
\label{sec: methods}


Based on real-world and simulated datasets, many DL-based 4D mmWave radar methods have been developed to enhance performance in adverse environments. As shown in Fig. \ref{fig: overall}, 4D mmWave radar-only methods achieve reliable and cost-efficient sensing through advanced preprocessing and feature extraction on 4D tensors or point clouds. In multi-modal methods, 4D mmWave radar compensates for camera or LiDAR limitations from adverse environments, such as noise and occlusion.

Different sensors exhibit varying sensitivities to different weather conditions. 
Therefore, corresponding to Fig. \ref{fig: overall}, some studies aim to improve general performance across multiple adverse conditions \cite{chae2024towards,chae2024lidar}, while others target specific conditions where LiDAR and cameras perform poorly through sensor fusion strategies \cite{huang2024l4dr,hahner2022lidar}.
In this section, we provide a comprehensive analysis of existing research, categorized by different challenging environments.


\newcolumntype{C}[1]{>{\centering\arraybackslash}p{#1}} 
\setlength\tabcolsep{5pt}
\begin{table*}[ht]
\centering
\caption{3D object detection methods on K-Radar dataset. AP$_{\text{3D}}$ values are in \%. IoU = 0.3. The best results are bold.}
\label{tab: main result vod}
\vspace{-1mm}
\begin{tabular}{C{2.2cm}|C{2.2cm}|C{1.2cm}| C{0.9cm}| C{0.9cm} C{0.9cm} C{0.9cm} C{0.9cm} C{0.9cm} C{0.9cm} C{0.9cm}}
\hline
Methods &Publication  &Modality &Total &Normal &Overcast &Fog &Rain &Sleet &Light Snow &Heavy Snow\\
\hline
RTNH \cite{paek2022k}&NeurIPS 2022 &R &47.4  &49.9  &56.7  &52.8  &42.0  &41.5 &50.6 &44.5 \\
\hdashline
PointPillars \cite{lang2019pointpillars} &CVPR 2019 &L  &45.4 &52.3 &56.0 &42.2 &44.5 &22.7 &40.6 &29.7 \\
VoxelNext \cite{chen2023voxelnext} &CVPR 2023 &L &67.7 &69.9 &74.5 &71.8 &66.3 &37.2 &83.3 &52.6 \\
Robo3D \cite{kong2023robo3d} &ICCV 2023 &L &56.4 &- &84.4 &50.0 &65.9 &25.0 &72.6 &44.4 \\
\hdashline
EchoFusion \cite{liu2023echoes} &NeurIPS 2023 &R+C &47.4 &51.5 &65.4 &55.0 &43.2 &14.2 &53.4 &40.2 \\
DPFT \cite{fent2024dpft} &IEEE TIV 2024 &R+C &56.1 &55.7 &59.4 &63.1 &49.0 &51.6 &50.5 &50.5 \\
\hdashline
InterFusion \cite{wang2022interfusion} &IROS 2022  &R+L &53.0 &51.1 &58.1 &80.9 &40.4 &23.0 &71.0 &55.2 \\
AttFuse \cite{xu2022opv2v} &ICRA 2022 &R+L &69.0 &66.8 &79.4 &88.6 &70.7 &59.2 &86.2 &58.6 \\
BEVFusion \cite{liu2023bevfusion}  &ICRA 2023 &R+L  &69.4  &69.9 &77.4 &66.8 &70.5 &43.2 &85.2 &61.6 \\
3D-LRF \cite{chae2024towards} &CVPR 2024 &R+L  &74.8 &81.2 &87.2 &86.1 &73.8 &49.5 &87.9 &67.2 \\
LOD-PRD \cite{chae2024lidar} &ECCV 2024 &R+L &73.2 &74.4 &76.9 &85.5 &70.3 &45.9 &87.8 &54.4 \\
L4DR \cite{huang2024l4dr}&AAAI 2025 &R+L &78.0 &77.7 &80.0 &88.6 &79.2 &60.1 &78.9 &51.9 \\
ASF \cite{paek2025availability} &Arxiv 2025   &R+L &87.3 &86.6 &89.8 &\textbf{90.7} &\textbf{88.6} &\textbf{80.0} &\textbf{88.8} &\textbf{77.5} \\
\hdashline
ASF \cite{paek2025availability}&Arxiv 2025 &R+C+L &\textbf{87.4} &\textbf{87.0} &\textbf{90.1} &\textbf{90.7} &88.2 &\textbf{80.0} &88.6 &77.4 \\
\hline
\end{tabular}
\vspace{-6mm}
\label{tab: k-radar}
\end{table*}

\subsection{Overall Adverse Environments}
Many studies propose generalized strategies to enhance robustness across multiple adverse environments. Although more datasets featuring extreme weather have been released, K-Radar \cite{paek2022k} remains the most widely used benchmark. Therefore, we first analyze methods for adverse conditions using K-Radar \cite{paek2022k}. Table \ref{tab: k-radar} summarizes 3D object detection approaches under adverse weather on K-Radar \cite{paek2022k}.

As the baseline for K-Radar \cite{paek2022k}, RTNH \cite{paek2022k} extracts radar feature maps from 4D tensors. RadarOcc \cite{ding2024radarocc} is the first study for 3D occupancy prediction using 4D mmWave radar under severe conditions. Doppler bins and the Top-\(N\) elements are encoded in each range to convert radar tensors into a sparse format.

Compared to 4D tensors, point clouds are more computationally efficient. When generating point clouds from 4D tensors, noise from adverse environments can be filtered out. RTNH+ \cite{10599871} designs a point cloud generating approach. Through CA-CFAR \cite{liu2024cfar} and sidelobe filtering techniques, noise from rain, snow, sleet, and fog is removed during the generation. Further hyperparameter tuning is conducted in \cite{paek2023enhanced}.
The domain shifts across different weather and road conditions are further examined in \cite{zhang2024exploring}, especially for snow, rain, and sleet weather. 

Radar-only methods often fall short of delivering satisfactory results due to their sparsity and noise. Hence, fusion methods are increasingly employed to improve sensing robustness.
EchoFusion \cite{liu2023echoes} generates radar BEV queries from 4D tensors and fuses the corresponding spectrum features with images. To address the degradation of cameras under adverse conditions, DPFT \cite{fent2024dpft} projects radar tensors onto both the front and BEV perspectives and fuses them with camera images.

4D mmWave radar-LiDAR fused methods integrate LiDAR's precise spatial information with 4D mmWave radar's robustness. With an attention mechanism, 3D-LRF \cite{chae2024towards} fuses each LiDAR voxel with its surrounding radar voxels. The weather-sensitive image features later serve as gating elements to control the fused information flow. 3D-LRF \cite{chae2024towards} highly increases the performance of methods without specifically considering adverse weather, such as InterFusion \cite{wang2022interfusion}, AttFuse \cite{xu2022opv2v}, and BEVFusion \cite{liu2023bevfusion}, especially for fog (19.3\%) and heavy snow (5.6\%) conditions.

However, extracting features from two modalities requires high computational costs. Distillation is often employed to transition from larger models to smaller ones. LOD-PRD \cite{chae2024lidar} trains a teacher model using the 4D mmWave radar-LiDAR fused method. Then, the weather-insensitive features are distilled into a LiDAR-only student model. 

Through bidirectional fusion, L4DR \cite{huang2024l4dr} mitigates LiDAR degradation in adverse conditions. 4D mmWave radar data compensates for missing LiDAR points via intra-modal feature extraction.
L4DR \cite{huang2024l4dr} surpasses the performance over LOD-PRD \cite{chae2024lidar}, especially under rain (8.9\%) and sleet (14.2\%) weather.

RTNH-AL \cite{10588669} introduces an auto-labeling approach for all-weather 4D mmWave radar tensors. A LiDAR-based detection network is trained and generates bounding box labels for radar tensors. Notably, performance improves when the model with generated labels is trained on a wider variety of weather conditions.

ASF \cite{paek2025availability} employs a canonical projection to enable consistency among 4D mmWave radar, LiDAR, and camera features. A cross-attention mechanism along patches enhances the robustness against sensor degradation and failure, achieving the best results in overcast, fog, and sleet weather.

Besides K-Radar-based methods, 4D RadarPR \cite{chen20254d} designed a context-aware 4D mmWave radar place recognition model for challenging scenarios from multiple datasets.
Doracamom \cite{zheng2025doracamom} fuses multi-view cameras and 4D mmWave radar for joint 3D object detection and semantic occupancy prediction, achieving the best results in night and rain on the OmniHD-Scenes dataset \cite{zheng2024omnihd}.

A human detection model proposed by \cite{skog2024human} processes multi-view radar heatmaps, including elevation, azimuth, range, and Doppler velocity dimensions. The data was collected in four challenging environments—an underground mine, a large car wash, an industrial tent, and an outdoor wooded area —where visibility is reduced by dust, water spray, and smoke. 
TransLoc4D \cite{10655664} is the first end-to-end 4D mmWave radar place recognition framework. After ego-velocity regression and RANSAC filtering \cite{shi2024ransac}, an attention mechanism captures local and global information of the point cloud, alleviating its sparsity and noise.

\subsection{Rain and Snow}

4D mmWave radar maintains strong robustness in rain and snow, while water on camera lenses can lead to blurred and defocused images.
In snowy conditions, cool temperature affects camera systems because of optical and mechanical disruptions. Reflected signals from rain and snow droplets create clutter for the LiDAR signal \cite{appiah2024object}.
Therefore, rain and snow severely degrade LiDAR and camera performance \cite{8249138}. To address this problem, some studies fuse 4D mmWave radar with cameras or LiDAR to specifically improve performance in rainy and snowy conditions.

A simulated 4D mmWave radar-camera tracking framework \cite{liu2024augmented} uses SUMO and CARLA software to simulate the highway roadside scenarios. It incorporates various rain conditions and demonstrates that the 4D mmWave radar-camera fusion method outperforms camera-only approaches during rainy nights.
Another multi-modal training strategy proposed by \cite{berens2024adaptive} combines real-world 4D mmWave radar and LiDAR data with simulated disturbances. Snow is modeled as adding points, losing points, and spatial shift on Astyx LiDAR data \cite{8904734}. Incorporating simulated snow noise can mitigate performance degradation. Similar effectiveness is demonstrated for fog on the VoD dataset \cite{apalffy2022}.

Compared to single-agent perception, cooperative perception enhances environment understanding by sharing observations among vehicles and infrastructure. However, multi-agent systems amplify environmental noise during communication. To address this, \cite{V2X-R} proposes a 4D mmWave radar-LiDAR fusion detection method for simulated V2X scenarios. Since LiDAR point clouds are degraded by weather noise, radar features are used to guide a diffusion process to denoise LiDAR data. The method is also tested on real-world datasets, showing better snow and rain weather improvement.

\subsection{Fog and Smoke}
Fog and smoke cause negligible effects on millimeter waves due to the large size disparity between their particles and wavelength \cite{berens2024adaptive}.
In contrast, fog and smoke had the largest impact on LiDAR because of the high extinction and backscattering coefficient \cite{8666747}. Besides, they can reduce the camera's image contrast. Thus, 4D mmWave radar can be used to compensate for the occlusions caused by fog and smoke in LiDAR and camera sensing.

Thus, some methods simulate fog on real-world LiDAR or camera data and fuse it with 4D mmWave radar data to increase robustness. 
Table \ref{tab: fog} shows the experiment results with different methods on the VoD-Fog dataset \cite{huang2024l4dr}.
TL-4DRCF \cite{zhang2024tl} implements a two-stage fusion of 4D mmWave radar and camera data. 4D mmWave radar point clouds are first projected onto camera coordinates. Later, voxels and pixels are correlated through frustum association and cross-attention mechanisms. It demonstrates superior robustness in foggy weather.
Similarly, L4DR \cite{huang2024l4dr} simulates different levels of fog on the VoD dataset \cite{apalffy2022}. The foggy noise for LiDAR is effectively filtered out by 4D mmWave radar during multi-stage fusion.

\setlength\tabcolsep{0.85pt}
\vspace{-3mm}
\begin{table}[h]
\centering
\caption{AP$_{\text{3D}}$ results (\%) on VoD-Fog dataset \cite{huang2024l4dr}.}
\vspace{-2mm}
\label{tab: modules}
\begin{tabular}{c|c|c|ccc|ccc}
\hline
\multirow{2}{*}{Method} &\multirow{2}{*}{Publish} & \multirow{2}{*}{\shortstack{Mod-\\ality}} & \multicolumn{3}{c|}{No Fog} & \multicolumn{3}{c}{With Fog} \\
\cline{4-9}
 & & & Car & Ped. & Cyc. & Car & Ped. & Cyc. \\
\hline
PointPillars\cite{lang2019pointpillars}\textsuperscript{\dag} &CVPR 2019 &L &73.5 &58.4 &79.0  &51.4 & 47.2 & 62.7 \\
\hdashline
TL-4DRCF\cite{zhang2024tl} &IEEE Sensors 2024 &R+C &43.7 &40.1 &64.2  &25.9  &23.2 &48.8 \\
\hdashline
InterFusion\cite{wang2022interfusion}\textsuperscript{\dag} &IROS 2022 &R+L &65.8 &70.1 &87.0  & 48.5 & 57.8 & 71.1\\
L4DR\cite{huang2024l4dr}\textsuperscript{\dag} &AAAI 2025 &R+L &76.6 &72.3 &90.4  &41.4 & 50.6 & 67.7 \\
\hline
\end{tabular}
\\[1pt]
\scriptsize{\textsuperscript{\dag} denotes results from \cite{huang2024l4dr} with fog level = 3. Others have no fog-level distinction.}
\label{tab: fog}
\vspace{-3mm}
\end{table}

Smoke conditions are simulated by deleting LiDAR scans in certain sequences in \cite{noh2024adaptive}. When LiDAR degradation is detected through radar-LiDAR overlap, static radar points will be used. Otherwise, static LiDAR points are used as input for odometry and dynamic point identification. Experiments show that 4D mmWave radar-LiDAR fusion achieves better odometry performance, while LiDAR-only methods suffer severe drift in dense smoke.

For depth estimation, traditional visible spectrum or near-infrared LiDAR often encounters issues such as noise and occlusion. RIDERS \cite{10623522} tackles these challenges by integrating 4D mmWave radar with infrared thermal camera data in challenging smoke-laden scenarios. The monocular depth predictions are aligned with radar points using a global scaling factor. And a transformer-based network is employed to estimate confidence in radar-pixel associations.

\subsection{Challenging Illumination}
Illumination has no impact on 4D mmWave radar. In contrast, low visibility conditions can significantly reduce cameras' field of view and image clarity, resulting in glare or dark shadows in images. Therefore, 4D mmWave radar is commonly used to fuse with the camera to compensate for the negative effects of poor lighting on images \cite{10422147}.

DADAN \cite{wang2025dadan} assigns higher weights to foreground points based on velocity and intensity, enabling accurate target detection even under strong illumination and foreground–background confusion.  
RCFusion \cite{10138035} fuses features from 4D mmWave radar and cameras within a unified BEV space, achieving high detection accuracy when image quality is degraded by low light or intense brightness.

Some methods categorize the TJ4DRadSet dataset \cite{9922539} into different lighting conditions: dark, dazzle, and normal.  
LXL \cite{10268601} introduces a sampling strategy using predicted depth maps and occupancy grids, showing that fusing 4D mmWave radar with images improves performance in dark scenarios due to informative vehicle lights, but suffers under excessive illumination.  
Following LXL \cite{10268601}, MSSF \cite{liu2024mssf} proposes a multistage sampling fusion network that maintains robustness and avoids performance drops in both dark and dazzle conditions.  
HGSFusion \cite{gu2025hgsfusion} further addresses direction-of-arrival estimation errors and uses modality synchronization to reduce the degradation of images, achieving better fusion performance under adverse lighting.

\section{Challenges and Frontiers}
\label{sec: discussion}

%


4D mmWave radar demonstrates strong potential in adverse environments for ITS due to its robustness. However, it faces challenges such as simulation realism, cooperative perception, and efficient noise handling. As its applications expand, future research will address these issues while exploring new tasks and use cases.

\textbf{Dataset simulation:}
Collecting data under real adverse weather is challenging, making simulation increasingly valuable due to its configurable environment settings. Some methods simulate entire scenes, while others augment real-world datasets with simulated weather effects like water droplets or spray \cite{huang2024l4dr}. 
However, the current simulation of 4D mmWave data often falls short of capturing complex real-world traffic. 
In future research, integrating meteorological principles into high-fidelity simulations will enable the generation of more realistic data. Moreover, capturing continuous data of the same scene from clear to adverse weather can support progressive robustness testing of algorithms.

\textbf{Cooperative perception:}
Beyond single-agent tasks, 4D mmWave radar has expanded to multi-agent systems \cite{V2X-R,yang2024v2x}. However, current V2X studies lack in-depth investigation of cooperative multi-modality feature-level fusion and face challenges such as communication-induced noise, synchronization, and time delay. 
Further exploration of adverse weather on communication among agents can provide more reliable benchmarks for cooperative perception.

\textbf{Motion-centric tasks:}
The Doppler velocity measurement enables 4D mmWave radar to excel in motion-centric tasks such as moving object segmentation, velocity prediction, and scene flow estimation in adverse conditions. Few existing methods have exploited velocity to denoise challenging weather. Future research utilizing velocity information can help models adapt to dynamic and complex traffic scenarios.

\textbf{Emerging use cases:}
When mounted on robots, 4D mmWave radar extends long-range perception capabilities. 
In dust- and smoke-filled mining environments, 4D mmWave radar has been deployed to detect static obstacles \cite{10684300}.
In the future, 4D mmWave radar will be further explored for new scenarios such as aquatic environments, construction sites, and human-centric applications.

\section{Conclusion}
\label{sec: conclusion}

This paper provides the first comprehensive review of 4D mmWave radar DL-based approaches to enhance robustness in adverse environments. The advantages of 4D mmWave radar over other sensors under challenging conditions are first highlighted. Then, we summarize the 4D mmWave radar datasets with various adverse weather and illumination scenarios, analyzing their characteristics and limitations. Based on open-source datasets, we analyze current 4D mmWave radar methods, including general strategies for multiple adverse conditions and specialized methods for specific conditions to compensate for other sensors' deficiencies. However, existing methods still face challenges such as data simulation and cooperative perception. We propose new directions for 4D mmWave radar development in adverse weather, including novel scenarios, tasks, and applications.
We hope this work provides valuable resources and inspires advancements for 4D mmWave radar in adverse environments.

\section*{Acknowledgment}
This research was conducted as part of the DELPHI project, which was funded by the European Union under grant agreement No 101104263. 





\bibliographystyle{IEEEtran} 
\bibliography{thebib} 

\end{document}